\documentclass[twocolumn]{article}
\usepackage{mwe}
\usepackage{authblk}
\usepackage[a4paper,top=2.54cm,bottom=3cm,left=2.54cm,right=2.54cm,marginparwidth=1.75cm]{geometry}
\usepackage{graphicx}
\usepackage{subfig}
\usepackage[font=small,labelfont=bf]{caption}
\usepackage{multirow}
\usepackage{booktabs}
\usepackage{mwe}
\usepackage{subfig}
\usepackage{amsmath}
\usepackage{authblk}

\usepackage[ruled,vlined]{algorithm2e}
\usepackage[utf8]{inputenc}
\title{\textbf{Automated Multi-Process CTC Detection using Deep Learning}}
\author{Authors: Elena Ivanova, Kam W. Leong, and Andrew F. Laine \thanks{Department of Biomedical Engineering, Columbia
University, New York, NY, USA Correspondence:
ei2169@columbia.edu}}
\date{}
\begin{document}

\maketitle

\begin{figure*}
  \includegraphics[width=\textwidth,height=7cm]{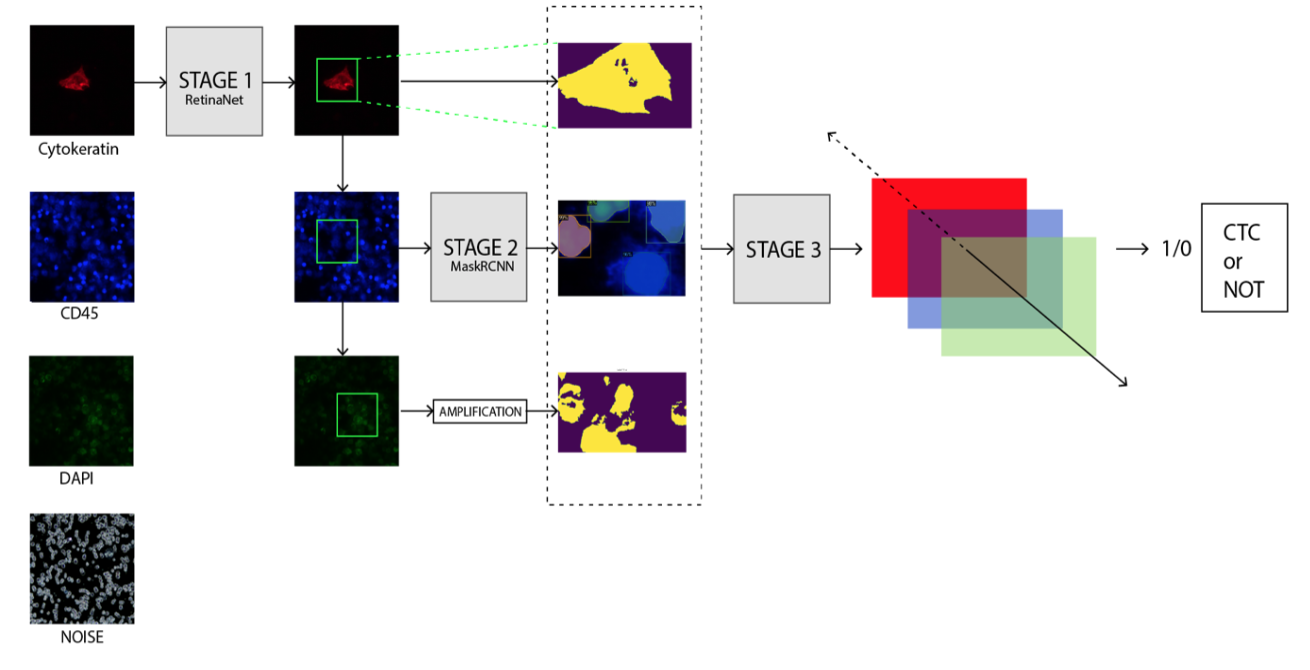}
  \caption{The pipeline is composed of 3 stages, Stage 1 identifying the Cytokeratin in the pictures, Stage 2 identifying the DAPIs and Stage 3 doing the final inference for the CTC.}
\end{figure*}

\begin{@twocolumnfalse}
\begin{abstract}
Circulating Tumor Cells (CTCs) bear great promise as biomarkers in tumor prognosis. However, the process of identification and later enumeration of CTCs require manual labor, which is error-prone and time-consuming. The recent developments in object detection via Deep Learning using Mask-RCNNs and wider availability of pre-trained models have enabled sensitive tasks with limited data of such to be tackled with unprecedented accuracy. In this report, we present a novel 3-stage detection model for automated identification of Circulating Tumor Cells in multi-channel darkfield microscopic images comprised of: RetinaNet based identification of Cytokeratin (CK) stains, Mask-RCNN based cell detection of DAPI cell nuclei and Otsu thresholding to detect CD-45s. The training dataset is composed of 46 high variance data points, with 10 Negative and 36 Positive data points. The test set is composed of 420 negative data points. The final accuracy of the pipeline is 98.81\%.
\end{abstract}
\end{@twocolumnfalse}

\subsection*{Introduction}
Circulating tumor cells (CTCs) are cells that intravasate into the vasculature from primary tumors and travel through the patient’s circulation \cite{Yang}. Often, CTCs extravasate the vasculature and seed the growth of metastases in distant tissues. While the mechanisms through which CTCs accomplish both intravasation and extravasation from the bloodstream are debated, it is generally accepted that earlier detection of CTC presence results in improved patient outcomes \cite{Heitzer}. Non-invasive liquid biopsies, which detect CTCs, frequently outperform tissue biopsies in assessing the disease progression and are generally becoming the preferred method for disease tracking \cite{Hofman}. 
Unfortunately, the concentrations of CTCs within the bloodstream in metastatic cancer patients are often extremely low by diagnostics standards (average prevalence of 1 CTC per billion blood cells \cite{Toner}) and hinder the ability to make swift treatment decisions. Quite simply, newer methods are required that can innovatively detect low concentrations of CTCs in order to win our centuries-long battle with cancer.
Determination of what circulating cell constitutes a CTC is no menial task, as the blood continually contains large cell populations – though most blood-present cells are small in comparison to CTCs. The standardized accepted criteria for identification of a CTC from a blood sample is that the cell must,
\begin{enumerate}
    \item[(i)] contain a nucleus
    \item[(ii)] exhibit a positive cytoplasmic expression of cytokeratin
    \item[(iii)] show a negative expression of CD45
    \item[(iv)] have a diameter larger than 5 µm \cite{Lorencio}
\end{enumerate} 
In order for a liquid biopsy sample to be deemed as CTC-positive, identification of multiple CTCs is frequently required. 

\subsection*{Background and Literature Review}

The number of Circulating Tumor Cells (CTCs) in blood is highly correlated with disease progression and tumor response to treatment. However, the detection of CTCs is very challenging because of their highly disproportional distribution relative to non-malignant cells in the bloodstream \cite{Miller}.

The traditional methods for circulating tumor cell detection still heavily rely on arduous manual process of a field expert needing to view multiple liquid biopsy images, and locate the suspicious malignant cells. Although, with the technological support gained initially from the rise of immunomagnetic CTC enrichment combined with flow cytometry \cite{Racila}, and later via multi-marker immunofluorescence staining \cite{Koudelakova}, manual identification of CTCs are a still time-consuming and error-prone process. 

Therefore, with the increasing interest towards CTC counting for prognosis and tracking the treatment progression, attention has turned to automating CTC detection methods \cite{Zhou}. Initial deterministic computer vision methods often use a grey scale image where each pixel has a grey scale value between 0 and 1. Two of the most popular of these methods - image segmentation and thresholding - generates object perimeters or segmentation masks, which are technically converting the object of desire on the picture to 1s while assigning 0 values for the pixels that do not contain the object of desire. Individual and combined uses of these methods have been prevalent in earlier cell segmentation tools such as CellProfiler\cite{cellprofiler}. The major disadvantage of such deterministic segmentation methods is that, they are particularly sensitive to noise and are unable to isolate individual cell markers when the objects are overlapping, or tightly adjacent to one another. Supervised traditional machine learning tools have offered some resolution to the above problems, primarily using probabilistic random forest based methods in conjunction with the deterministic segmentation procedures to more accurately detect and isolate cells, and for subsequent classification \cite{Carpenter}. Traditional machine learning methods are still dependent on selection of factors via deterministic segmentation methods; thus even when probabilistically weighed, they carry forward the bias from the deterministic methods mentioned above.

Advancements in deep learning have removed the need for deterministic selection of features, and have thus removed the inherent biases and limitations from feature creation and selection. This allowed immense increases in accuracy and generalization in image classification \cite{Krizhevsky} and later image segmentation. Convolutional Neural Networks (CNNs) were used to identify object masks in a pixel-level classification task\cite{Sermanet} through the spatial information inherent in image data. Nonetheless, segmentation via CNNs is an inefficient process, and requires massive amounts of data to reach high accuracy, yet they still were effectively used in both cancer cell detection \cite{Xing} and even in CTC detection. These problems were largely alleviated by advancements in two fronts. From a model advancement perspective, the developments in region proposal networks and single-shot detection algorithms, like Yolo \cite{YOLO} and RCNN \cite{faster-rcnn} offered immense gains in speed and accuracy, yet were limited to bounding box level segmentation, which with the introduction of Mask-RCNN's \cite{Mask-rcnn} finally allowed the ability to segment the object perimeters with precision. From a transfer learning perspective, with creation of enormous datasets like ImageNet and COCO, and hardware improvements that allowed extreme improvements in training speeds, it was shown that using large pretrained models and fine-tuning the models on specific tasks in the image classification and segmentation space, would yield incredible increases in accuracy \cite{imagenet}. This created an opportunity to tackle tasks in a variety of fields, where previously the lack of annotated data had made deep learning applications inaccessible. On account of this, the field of CTC detection also saw recent attempts to use pretrained  Faster-RCNN models for more accurate detection of CTCs [19, 20, 21]. However, these were attempted to be applied directly on darkfield phase contrast images, and as far as we are aware of never before comprehensively on multi-channel stain enriched microscopic images.

Most of these attempts have aimed to perform detection directly on multi-channel enriched microscopic images \cite{He, Zhang, Ciurte, Van}. This limits the amount of signal provided into the models to only differences in shape between the cells. However, when the data is augmented with separable channels and layers in a single microscopic snapshot of the bloodstream, it enables the decision criteria to be based not only on the shapes of the objects but also the percentage of overlap across layers to accurately detect the precise cell of the CTC, where each layer corresponds to a Cytokeratin (CK) stain, DAPI markers and CD-45s. 

The challenging constituent of this task from an artificial intelligence standpoint is not to determine the extent of overlap between the markers from each channel, but to identify the "proper" objects in each channel and distinguish them from noise. In order to establish, by the decision criteria given below, whether a CTC exists in a given multi-channel set of images, we need to accurately perform each of the following tasks
\begin{enumerate}
    \item[(1)] Determine the area of interest around the Cytokeratin stain
    \item[(2)] Identify the boundaries for the cell nucleus marked in the DAPI layer
    \item[(3)] Identify the boundaries for the CD-45 enzyme
    \item[(Final)] Apply the decision criteria, to classify each identified cell as CTC or not
\end{enumerate}

The entire pipeline can then be represented denoting each model for each stage, $f_1, f_2, f_3, D$, as
\[D \circ f_3 \circ f_2 \circ f_1(X)\]
where X represents the multi-channel input data, and $D$ the final decision layer, based on the above decision criteria.

There are two immediate advantages that come with the availability of multi-channel data. First, it permits the overlapping objects of CD-45, CK stain and DAPI markers to be immediately separable (yet, not individually identifiable). Second, this allows us to focus the detection task to a specific region in the image marked by the CK stain, reducing the noise coming from the multitude of DAPI markers in a given microscopic snapshot. In contrast to previous methods, we only attend to the DAPI markers overlapping with the CK stain rather than attempting to perform classification on each individual DAPI marker \cite{Mao} or solely on the holistic view of the microscopic snapshot \cite{Ciurte}.

\subsection*{Our Approach}
The particular case in hand is unique due to the size of the dataset: 46 data points with 36 Positive and 10 Negative samples.
Given these limitations, we developed a 3-stage solution that optimizes for over-fitting to the 46 data points, while also being robust enough to handle the variance in the real-life data. The pipeline is composed of 3 separate stages, 2 with an intelligent deep learning model and the last stage being a purely probabilistic model. We argue that if a larger task is broken down to its sub-tasks, and such a break down happens by splitting data points to a higher detail form (making a single input 3 layers), the total bias in the models dedicated to the sub-tasks will be less than the total bias of a single model trying to solve for the problem. 
We refrained from using any augmentation to not introduce any synthetic bias to the real-life problem. The segmentation of the problem to it's sub-problems allows for a high success rate without the necessity to create more data points.

Stage 1 of the pipeline is Cytokeratin Localization, where we identify the Cytokeratin regions in the multi-channel enriched images. The pipeline here takes an enriched Cytokeratin Layer as input (Figure 2), and outputs coordinates of every Cytokeratins bounding box in the input picture.

\begin{figure}[h]
    \centering
    \includegraphics[width=7cm]{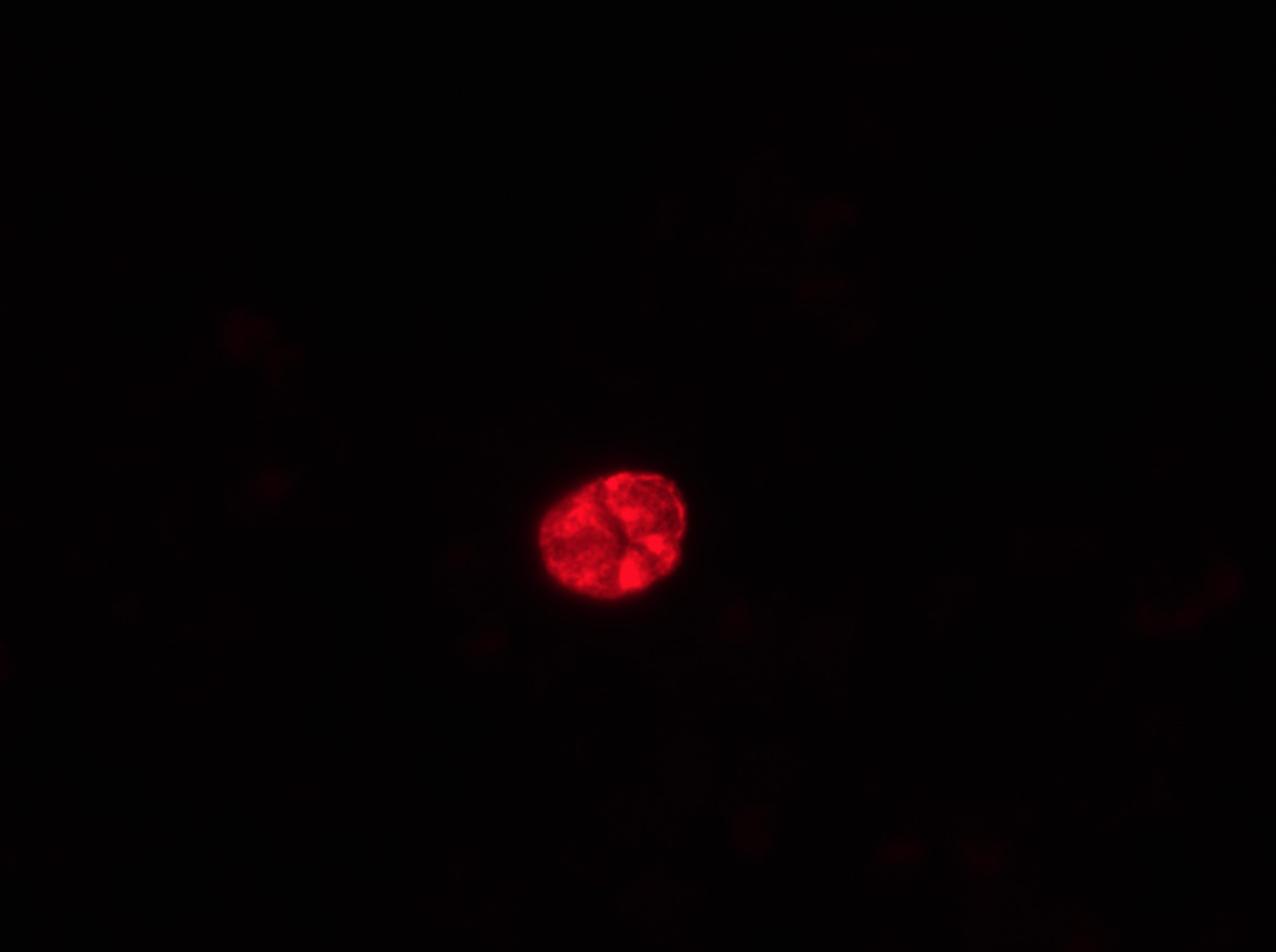}
    \caption{Cytokeratin Stain Colored Red in a Black-Field Microscopic Image}
    \label{fig:galaxy}
\end{figure}

By definition, for a DAPI to be classified as a CTC, it needs to be overlapping with a Cytokeratin, and it should not have any CD-45 underlying it. Therefore Cytokeratins play a critical role in the decision process. Through the identification, we localize and limit the search problem to only the Cytokeratin region. 

\begin{figure}[h]
    \centering
    \includegraphics[width=4cm]{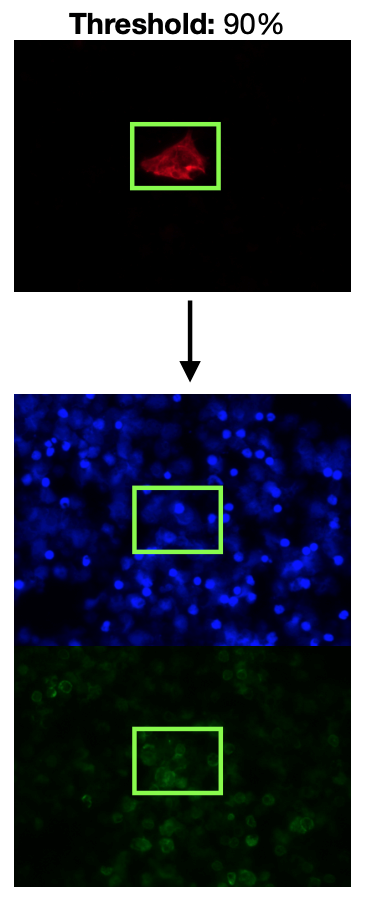}
    \caption{Stage 1: Localized Bounding Boxing Pivoted Around the Identified Cytokeratin}
    \label{fig:galaxy}
\end{figure}

The algorithm at this stage uses a RetinaNet architecture pretrained on the ImageNet and COCO datasets \footnote{ImageNet and COCO datasets are a collection random, everyday object and have 1.3 million and 330,000 training data points respectively. While ImageNet offer 1,000 different classes, COCO is more specialized in only 80.} and fine-tuned on the Cytokeratin data to draw a bounding box around the Cytokeratin in the picture. 

Once successfully localized (Figure 3), the problem statement moves on to identifying DAPIs within the localized region. Therefore, in the next step of the pipeline, DAPI Detection, we detect the DAPIs in the Cytokeratin region. Stage 2 takes in the DAPI layer cropped to match the bounding box for the Cytokeratin as the input and outputs the coordinates for the DAPI masks in the given region. The cropping serves to localize the problem. 

\begin{figure}[h]
    \centering
    \includegraphics[width=7cm]{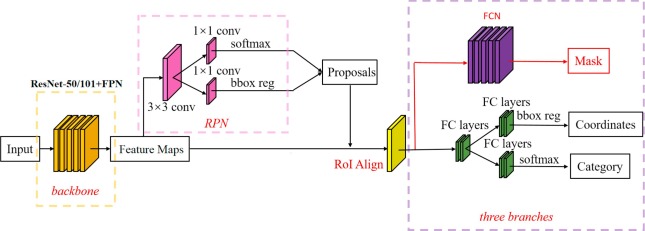}
    \caption{Stage 2: Mask-RCNN Architecture}
    \label{fig:galaxy}
\end{figure}

This layer uses a Mask-RCNN to identify the DAPIs within the region of interest (Figure 5). The data is cleaned through transforming the localized images to B/W in this stage. Then using a Mask-RCNN pretrained on the COCO dataset and finetuned on the DAPI data, we detect the DAPIs within the localized region (Figure 5). The training of the transfer learning model comes with a traditional train/test 0.8/0.2 split, focusing on the detection of the DAPIs with a mAP optimization function. The threshold of detection used for identifying the DAPI in the final output is 90\%.

\begin{figure}[h]
    \centering
    \includegraphics[width=4cm]{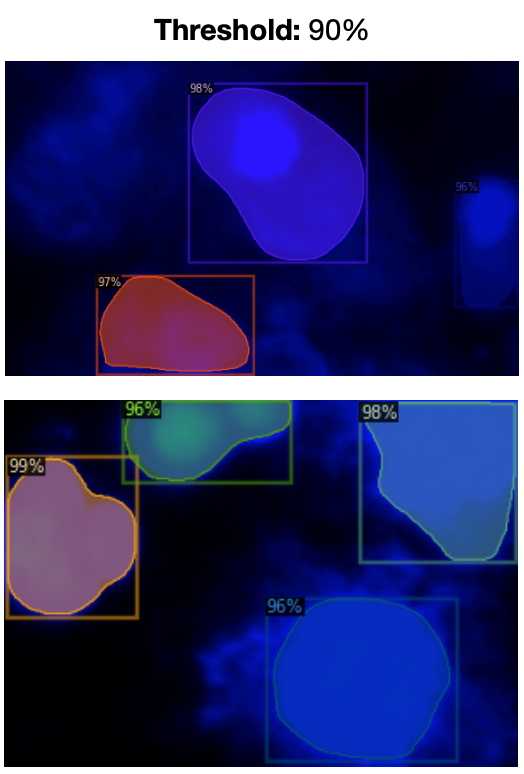}
    \caption{Stage 2: Mask-RCNN Masks for DAPIs with their confidences}
    \label{fig:galaxy}
\end{figure}

Once the DAPI identification and Cytokeratin identification are complete, we move on to the final decision stage to determine the overlap.

Stage 3 takes in Cytokeratin Cropped Images (from Stage 1), CD-45 Image cropped according to the Cytokeratin bounding box, DAPI mask within the bounding box (from Stage 2) and outputs bounding boxes around the CTC nuclei, and the confidence value of the bounding box.

At stage 3, we check if there is a CD-45 enzyme attached to these CTC candidates. In order to do this, we first need to identify CD45 enzymes in the corresponding layer. Due to the faint and uniform nature of the CD45 objects, we identify the area of each individual enzyme by Otsu thresholding with a dynamic threshold $t$ mapping each pixel to a Black or white pixel, i.e.

\[o(p) = \begin{cases} 255\ \text{(white)} & \text{if } p > t \\ 0 \ \text{(black)} & \text{otherwise} \end{cases}\]

where $t$ is optimized per image to account for varying levels of faintness and noise in each CD45 layer to amplify the data signal.

\begin{figure}[ht]%
 \centering
 \subfloat[Mask for Cytokeratin Stain]{\includegraphics[width=0.2\textwidth]{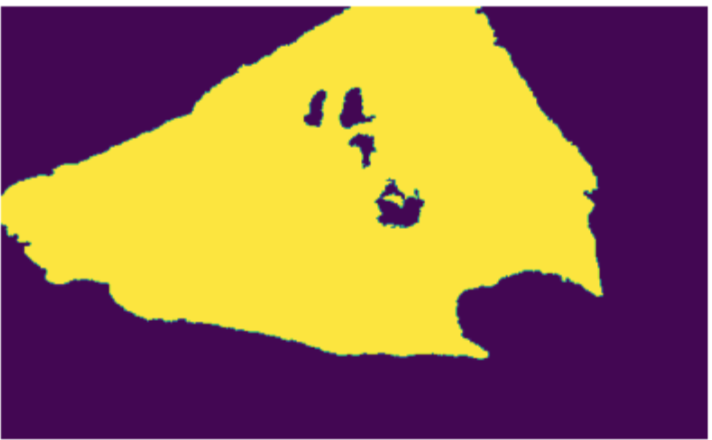}\label{fig:a}}%
 \subfloat[Mask for DAPI cells]{\includegraphics[width=0.2\textwidth]{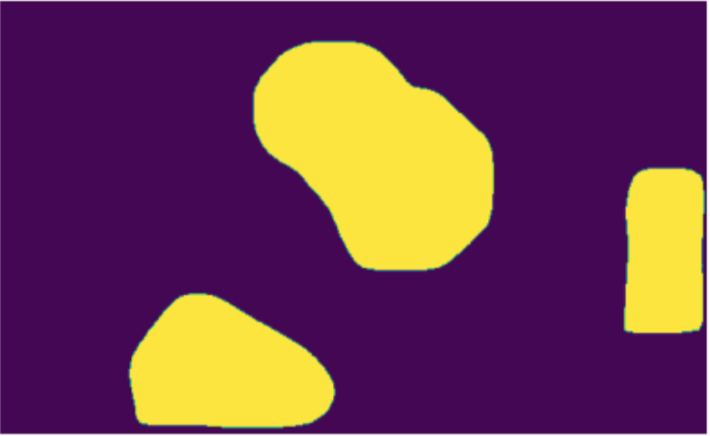}\label{fig:b}}\\
 \centering
 \subfloat[Mask for CD-45 via thresholding]{\includegraphics[width=0.25\textwidth]{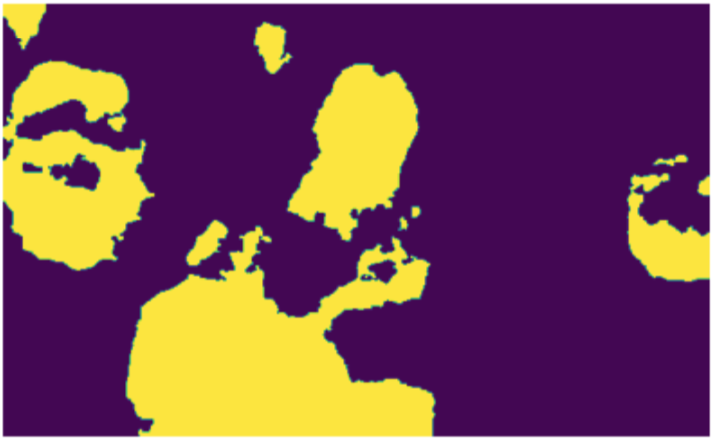}\label{fig:c}}%
 \caption{Each cell is identified from the DAPI layer masks gathered from stage 2, as represented in (b), and overlaps are checked against Cytokeratin region gathered from Stage 1, as represented in (b) and CD-45 gathered via thresholding as described above, represented in (c).}%
 \label{some-label}%
\end{figure}

Having thus identified the area of CD45 enzymes, we finally check percentage of overlap with each CTC candidate cell nucleus from Stage 1 and 2, with (i) the Cytokeratin stain and (ii) the CD45 enzyme. In order to maximize the precision-recall we define two parameters for each step of the decision criteria:
\begin{align*}
    r_1 &= 0.17 \\
    r_2 &= 0.2
\end{align*}

where $r_1$ is the minimum overlap percentage with a CTC candidate mask (DAPI) and Cytokeratin Stain, and $r_2$ is the minimum overlap percentage with a CTC candidate mask and CD45 enzyme, optimized via grid search such that for a candidate cell C,
\begin{align*}
    \underset{r_1, r_2}{\arg\max}\  p(C = \text{CTC} | C = \text{DAPI};r_1, r_2)
\end{align*}

\subsection*{Results}
The final pipeline was tested on 420 negative samples. Out of 420, we predicted 415 as negative and 5 as positive. Giving us a net accuracy of 98.90\%. Out of 415, the pipeline couldn't detect a Cytokeratin in 130 images, couldn't detect DAPIs after predicting Cytokeratin in 170 images and executed the full pipeline on 120 images. This gives us 5 wrong predictions in the 120 fully evaluated pictures, which makes the final step of the pipeline 95.8\% accurate. 

\subsection*{Discussion and Conclusions}
It is valuable for us to review the incorrect predictions to interpret the potential short-comings of the pipeline. Here are several examples of the inaccurate identifications:

\begin{figure}[htp]
    \centering
    \includegraphics[width=6cm]{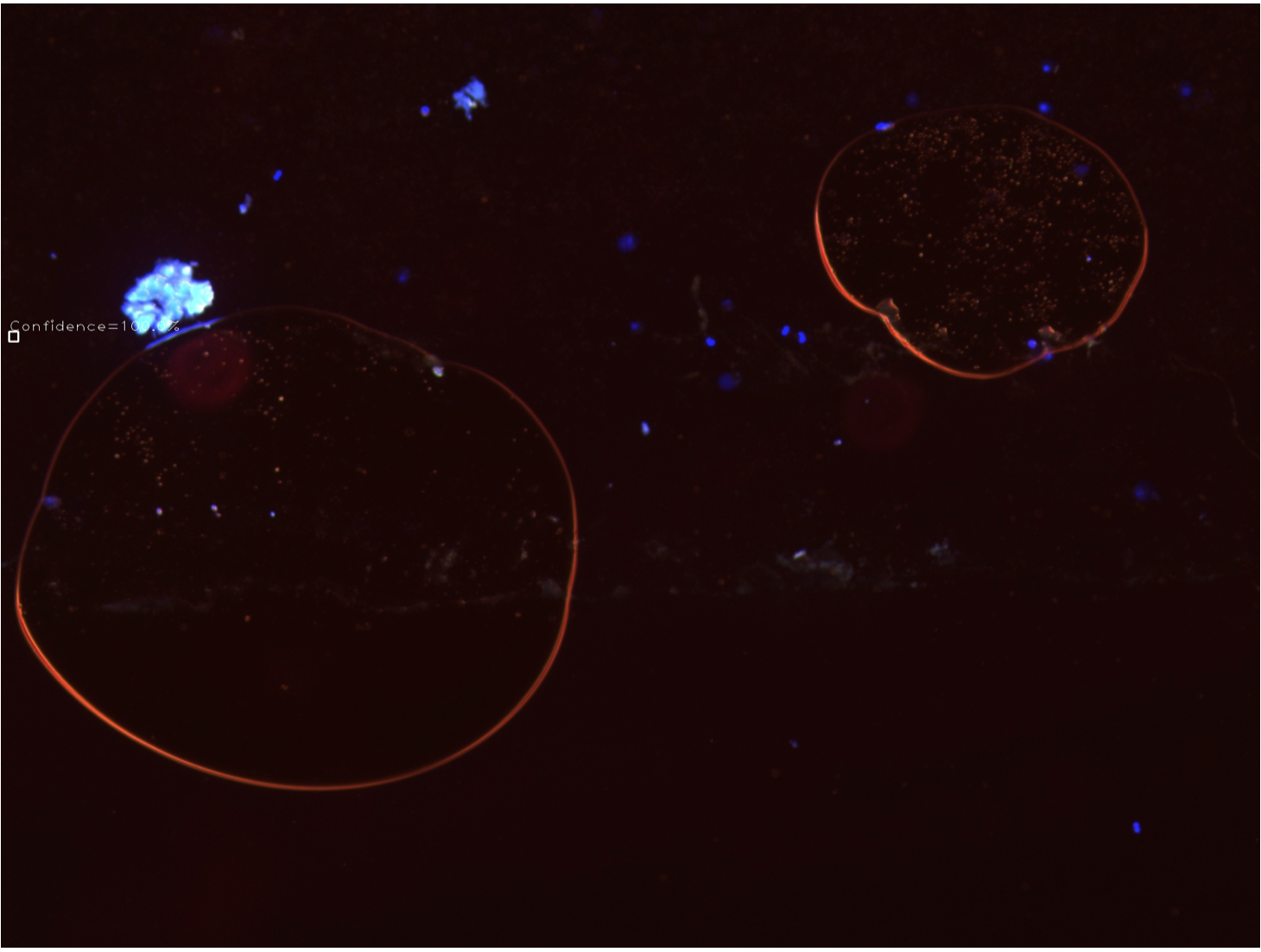}
    \caption{Incorrect Prediction 1: confidence of 100\%.}
    \label{fig:galaxy}
\end{figure}
\begin{figure}[htp]
    \centering
    \includegraphics[width=6cm]{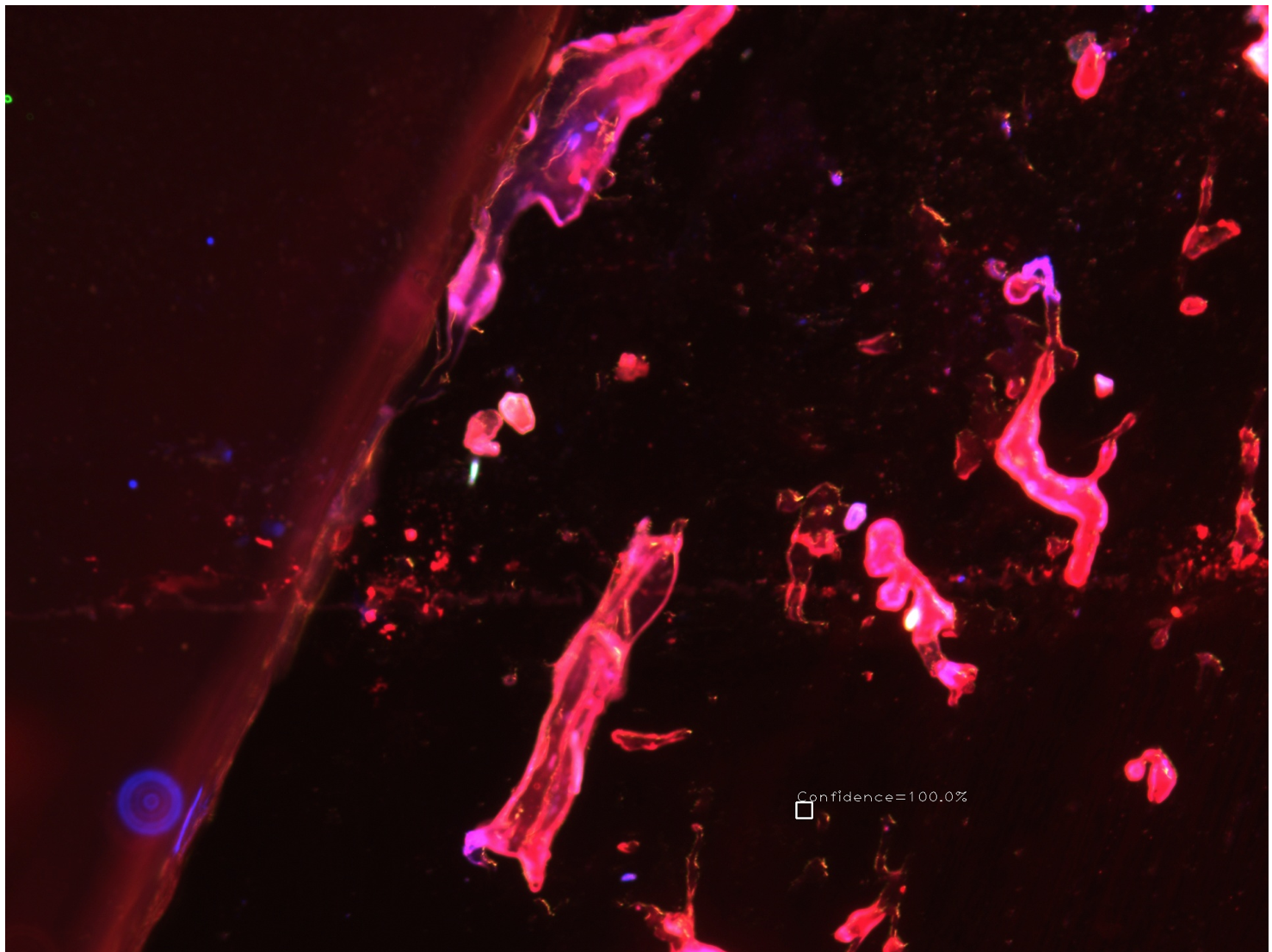}
    \caption{Incorrect Prediction 2: confidence of 100\%.}
    \label{fig:galaxy}
\end{figure}
\begin{figure}[htp]
    \centering
    \includegraphics[width=6cm]{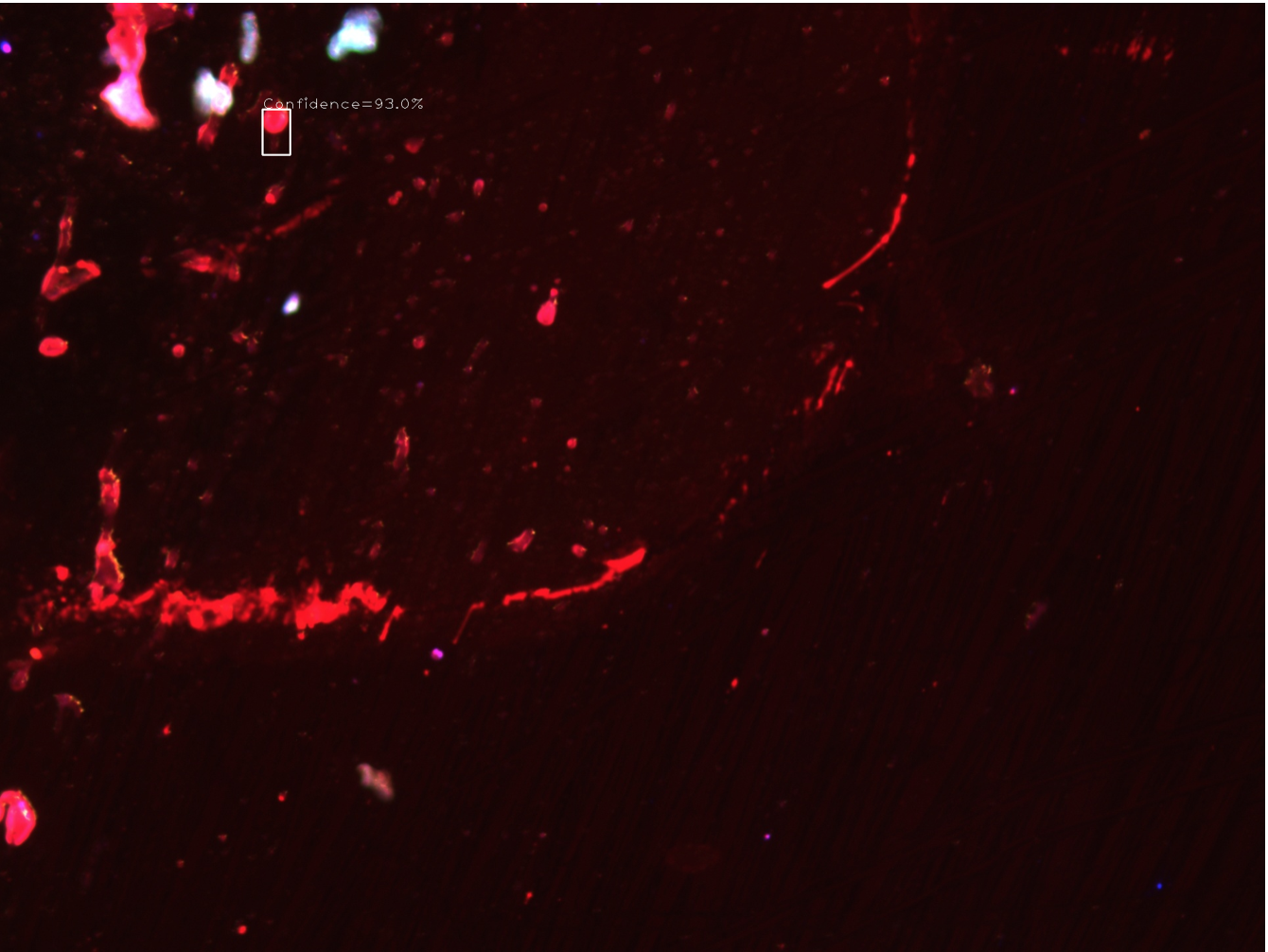}
    \caption{Incorrect Prediction 3: confidence of 93\%.}
    \label{fig:galaxy}
\end{figure}

Given the consistently small bounding boxes in all of these images (Figure 7-9), we hypothesise that the mistakes can and should be attributed to noise variance in the train and test datasets. The train dataset had cleaner multi-channel enriched images with no noise, whereas in the test dataset the provided images were noisy and incorrectly taken. Figure 9 is an example of an erroneous picture, where chemical stains from flawed photography are visible. 

Another important point is the variance we observe in the training data versus the test data, and how different the two batches are from one another. The train data as seen in Figure 10, is very clean non-noisy images, with visible round Cytokeratins found in the center region of the layer. Compared to its counterparts in the test data (Figure 11), the noise present in the test samples, and the variance of the noise with inconsistent Cytokeratin shapes and sizes, coupled with the camera flares (Figure 11 (b)), introduces a completely novel task to our pipeline. We are aware of the fact that the industry implementation of the pipeline will benefit from clean inputs like in the training sample, and attribute the erroneous classifications to the variance in the two datasets. We observe the same variance in shape and noise in data for the other layers provided for testing. 

While performing well on the high variance dataset is indicative of the robustness of the 3-stage model, it also serves as proof of the room available for improvement. We can focus on a handful of issues based on the test results: ranging from data cleaning and denoising, to re-training and variance handling.

\begin{figure}[h!]
\begin{tabular}{cc}
\subfloat[Train Data Sample 1]{\includegraphics[width = 1.3in]{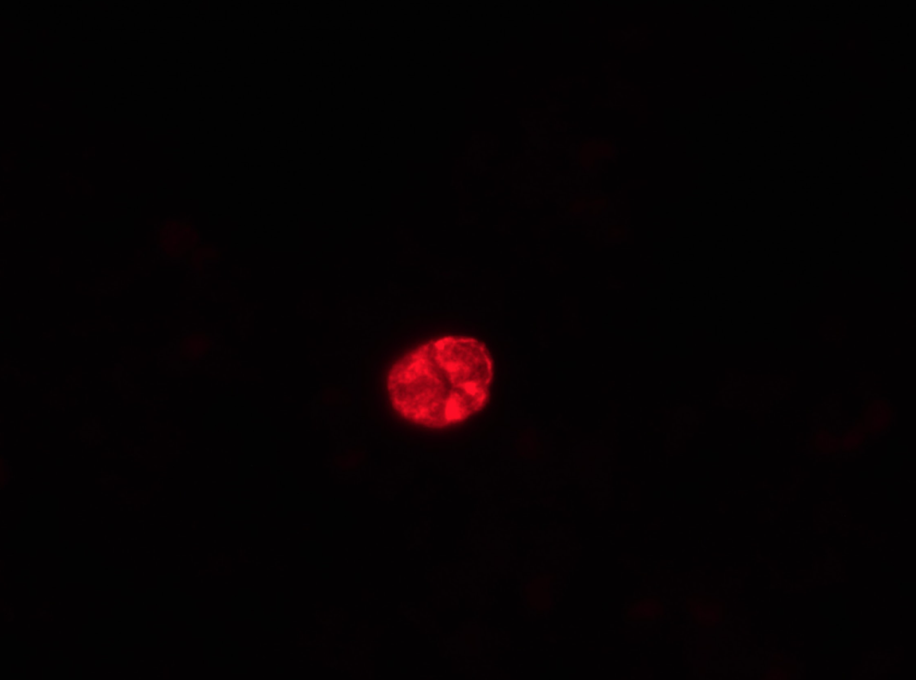}} &
\subfloat[Train Data Sample 2]{\includegraphics[width = 1.3in]{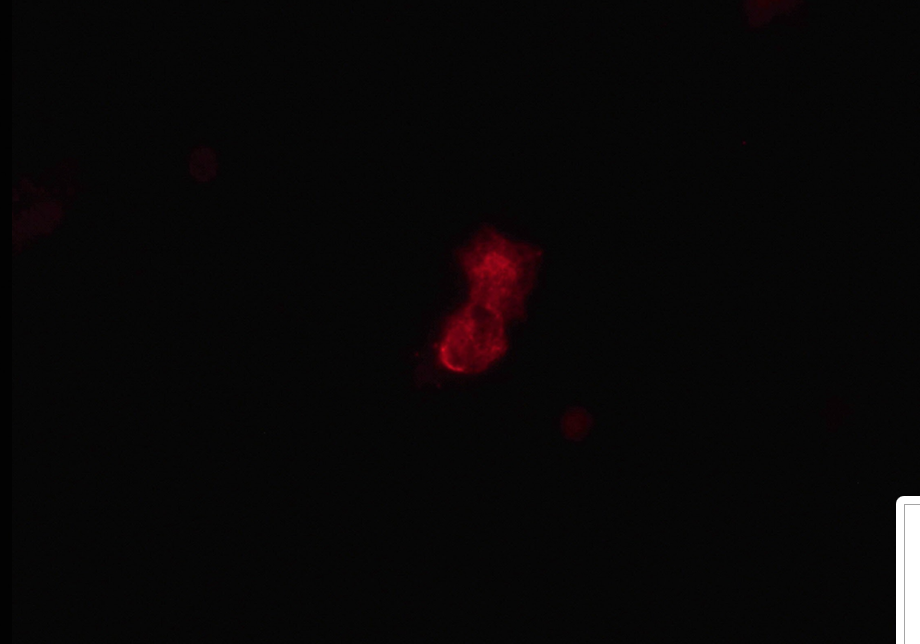}} \\
\subfloat[Train Data Sample 3]{\includegraphics[width = 1.3in]{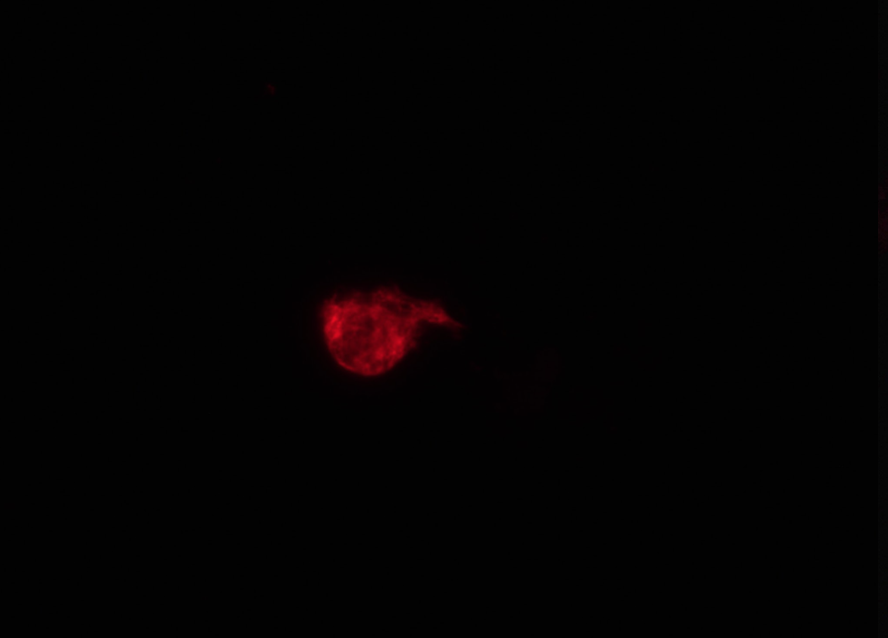}} &
\subfloat[Train Data Sample 4]{\includegraphics[width = 1.3in]{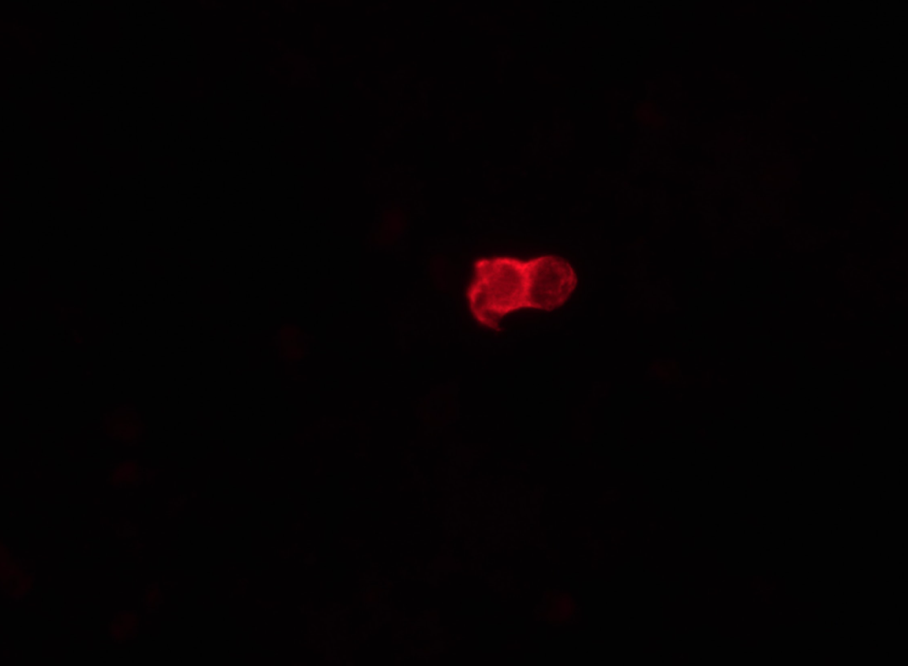}} \\
\subfloat[Train Data Sample 5]{\includegraphics[width = 1.3in]{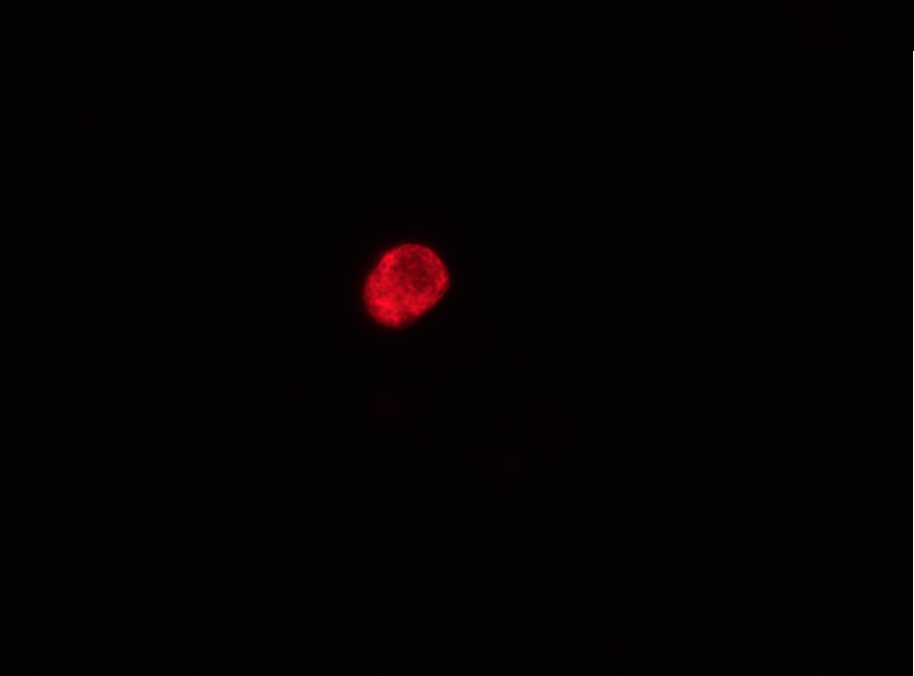}} &
\subfloat[Train Data Sample 6]{\includegraphics[width = 1.3in]{Screen_Shot_2021-08-15_at_6.10.20_PM.png}}
\end{tabular}
\caption{Train Data Sample for the Cytokeratin Multi-channel Enriched Images}
\end{figure}

\begin{figure}[h!]
\begin{tabular}{cc}
\subfloat[Test Data Sample 1]{\includegraphics[width = 1.3in]{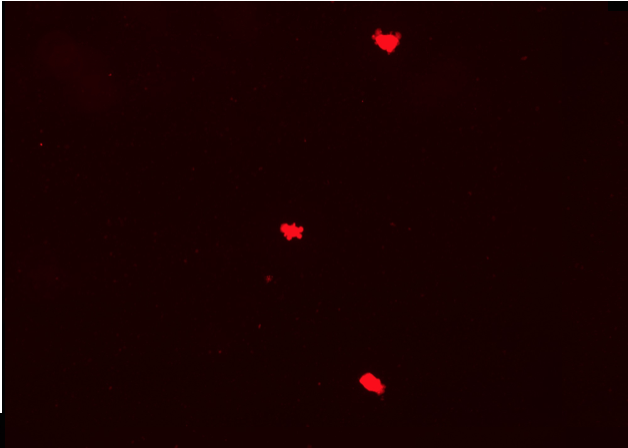}} &
\subfloat[Test Data Sample 2]{\includegraphics[width = 1.3in]{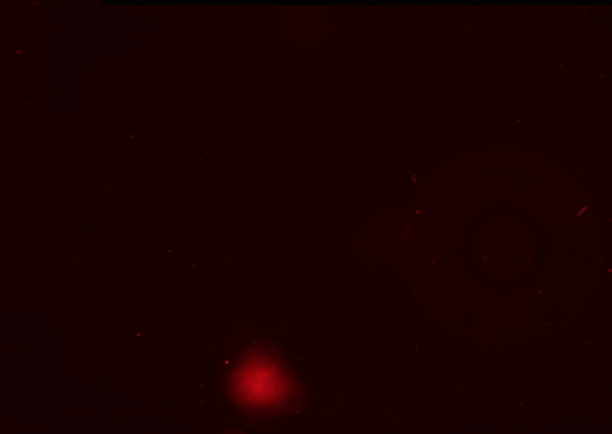}} \\
\subfloat[Test Data Sample 3]{\includegraphics[width = 1.3in]{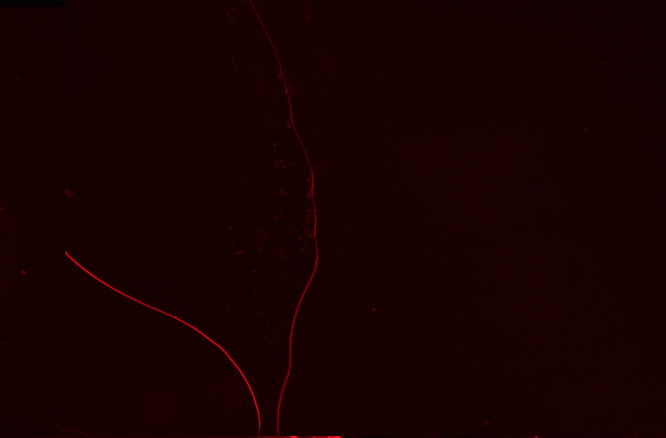}} &
\subfloat[Test Data Sample 4]{\includegraphics[width = 1.3in]{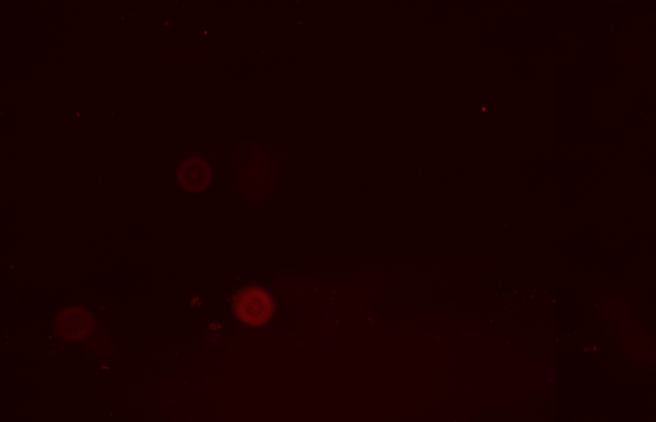}} \\
\subfloat[Test Data Sample 5]{\includegraphics[width = 1.3in]{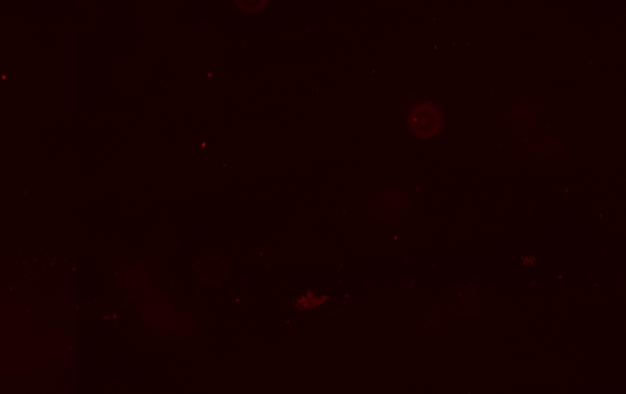}} &
\subfloat[Test Data Sample 6]{\includegraphics[width = 1.3in]{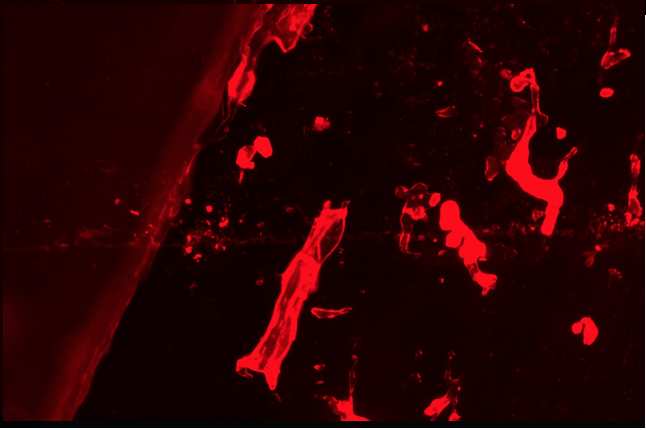}}
\end{tabular}
\caption{Test Data Sample for the Cytokeratin Multi-channel Enriched Images}
\end{figure}

Furthermore it is important to discuss the imbalance present in the testing dataset, which mirrors the real-life data. The training dataset (Figure 12) is composed of 78\% CTC positive images. Whereas the testing dataset is composed of 420 Non-CTC samples. Given that CTCs are very rare in real life blood samples, the test dataset is more representative of the distribution of CTCs the pipeline will face in the real world. However, by training the pipeline on a high positive CTC sample dataset, we managed to avoid the dominant class bias in the pipeline, where the pipeline predicts a certain class for all inputs since almost always the input belongs to that dominant class. 

\begin{figure}[h!]
    \centering
    \includegraphics[width=4cm]{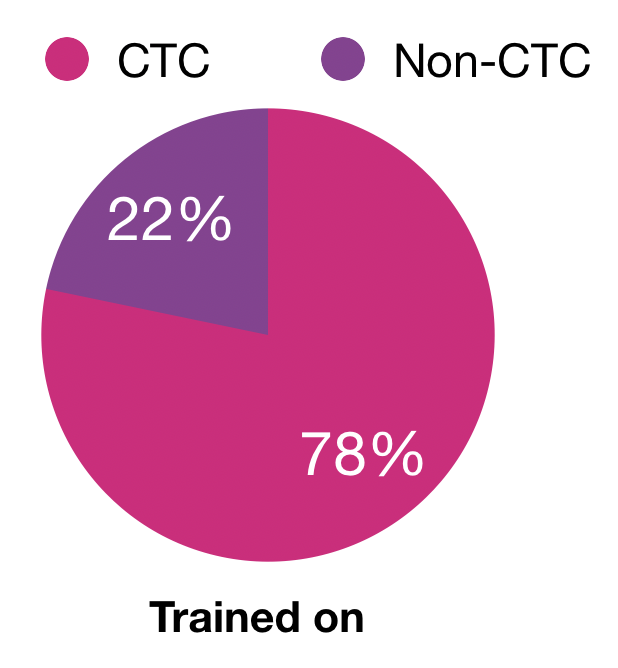}
    \caption{Training Data Balance}
    \label{fig:galaxy}
\end{figure}

Furthermore a point of concern for us was the bias introduced by the pipeline. When detecting CTCs, a false positive is significantly less fatal for the patient, than a false negative. As such, we wanted to make the pipeline predict the sample as positive when ever in doubt. The test on 420 negative data samples with 5 false positives works to prove the introduced bias' effect in the pipeline. 

When we look at the pipeline in terms of how the data flows through it, we first see the Cytokeratin detection layer. This layer failed to detect 130 at detecting 130 layers' Cytokertins. There needs to be a further investigation into the validity of the negative predictions such that some layers actually didn't have Cytokeratin. Furthermore the same argument applies to the negative classifications in Stage 2. While there Stages do present the possibility to act unexpectedly due to their low data points, this only proves the improvability of the pipeline with basic training. 

While the one-shot detection models would have also performed well on the detection process, the case we had in our hands had unique limitation that impeded with such deep models' performances. The 3-stage model solution allowed us to utilize the low number of data in the most effective way while giving us the opportunity to isolate the critical layers and amplify their signals - especially in the case of CD-45s. 

In conclusion, we observe a powerful performance by the pipeline in identifying CTCs apart from the noise and CTC similar substances. In the results we observe a great ability to identify non-CTCs with a very high accuracy. Unlike any known research in the field, the multi-modal architecture of the pipeline uses Mask-RCNN and RetinaNet, two state-of-the-art deep learning models trained on the exhaustive COCO and ImageNet datasets. The success of the pipeline attests to the power of transfer learning, a strategic approach to small dataset problems like ours. While the model has achieved high accuracy in a stacked model pipeline like ours, its small training dataset is its greatest weakness. The model would benefit greatly from more data points, as well as, an exhaustive test dataset to deeply understand its extensive functionality. Our post-hoc analysis of final results points at the strength of the model in it's Stage 3 inference, and the robustness of the model in identifying Cytokeratin/DAPIs from their respective layers. In its deployment to a industrial process, the pipeline will be accompanied by a live-feed data recursion software that will incrementally train the model post-manual annotation of the incoming training data. Such a process could also be supported by a semi-supervised annotation algorithm, which would introduce further efficiency into the software. The pipeline stands as the successful first version, that will grow rapidly upon the retrieval of new data. 

\subsection*{Code and Dataset Availability}
The code and documentation will be available on GitHub on article publication.

\subsection*{Acknowledgements}
We thank Prof. Andrew Laine and Prof. Kam Leong (Columbia University, Department of Biomedical Engineering) for comments, assistance and support. We gratefully acknowledges the
value of the data from the Strandsmart Inc.'s RD team and Dr. Mark Connelly.

\subsection*{Conflicts of interest}
Elena Ivanova has financial interest in Strandsmart Inc.

\subsection*{Methodology}
\subsubsection*{Confidence Metric}
The final confidence is an aggregate measure based on the three metrics: (i) the probability that the identified Cytokeratin region, actually contains the Cytokeratin object, $p(CK)$ (ii) the probability that the identified mask for the DAPI contains the nucleus, $p(C)$ (iii) product of

\begin{enumerate}
    \item[a.] percentage of overlap between the Cytokeratin and DAPI cell nucleus computed by,
    \[p(CK | C) = \frac{A(\text{CK}\cap\text{C})}{A(C)}\]
    
    \item[b.] percentage of overlap between identified CD45 mask and DAPI cell nucleus, computed by
    \[p(CD45 | C) =\frac{A(\text{CD45}\cap\text{C})}{A(C)}\]
\end{enumerate}

where $A(x)$ is the count of non-zero pixels for the object mask $x$, $CD45, CK, C$ are the masked areas, where the pixel value equals 1 if the pixel location falls in the masked area, and 0 otherwise, respectively for the CD45 mask (computed via thresholding), CK mask (computed via thresholding for the area identified by the RetinaNet), and DAPI mask (identified by the Mask-RCNN). Finally the confidence score is defined as the product of the above metrics,
\[ p(CD45|C)p(CK|C)p(C)p(CK)p(CD45)\]
with the assumption that p(CD45) = 1, if CD45 is identified via thresholding.
 
\subsubsection*{Decision Criteria}
Final decision criteria represented as
\begin{align*}
    \text{if}&\ \frac{A(\text{CK}\cap\text{C})}{A(C)} > r_1, \text{and} \\
    \text{if}&\ \frac{A(\text{CD-45}\cap\text{C})}{A(C)} > r_2 \\
    &\text{then}\ C \text{ is a CTC.}
\end{align*}

\noindent where $A(CK \cap C)$ denotes the overlapping area of the candidate cell (from Stage 2) and the Cytokeratin stain (from Stage 1), and $A(CD45 \cap C)$ denotes the overlapping area of the cell with the CD45 enzyme mask (Stage 3).

\begin{algorithm}
\SetAlgoLined
\KwResult{$\text{CTC} \rightarrow [0,1]$}
 $r_{1} = 0.17, 1 \geq r_{1} \geq 0$;\\
 $r_{2} = 0.2, 1 \geq r_{2} \geq 0$ \\
\eIf{$\|DAPI\| > 0, \|CK\| > 0$}{\eIf{$A(\text{DAPI}\cap\text{CK}) /A(C)<r_{1}$}{\text{return 0}}{\eIf{$A(\text{DAPI}\cap\text{CD-45})/A(C)>r_{2}$}{\text{return 1}}{\text{return 0}}}}{\text{return 0}}
\caption{Psuedo-code for Stage 3}
\end{algorithm}

\subsubsection*{Transfer Learning}
Due to the small training data (46 samples - 36 Positive, 10 Negative) we chose to use pre-trained neural networks, that were fine-tuned on our data, by training the last few layers of the models. This allowed for smarter inference, due to the leveraging of information from the millions of data points the models had seen, and also allowed us to combat potential over-fitting of the models. 

\subsubsection*{Retina Net}
Retina Net is a complex deep learning model that deals with computer vision problems. As opposed to most of its peer networks, it is composed of a main network and two specific to task sub-networks. The main network is tasked with generating the convolutional feature map over the whole input picture. The two sub-nets are tasked with carrying forward a object classification task in the local region, and draw the bounding boxes on the identified object of interest. This makes RetinaNet powerful in small scale identification, with single-stage architecture. RetinaNet's innovative success also comes from the focal loss it uses - applying "a modulating term to the cross entropy loss in order to focus learning on hard negative examples." \footnote{https://paperswithcode.com/method/retinanet} The configuration for our RetinaNet is:
\begin{enumerate}
    \item Batch Size: 16
    \item Base Learning Rate: 0.01
    \item Steps: (60000, 80000)
    \item Max Iter: 90000
    \item Number of Residual blocks [depth]: 50
\end{enumerate}

The main reason why we didn't use a Mask-RCNN to mask the Cytokeratin in this stage is to make sure that there are no partial DAPIs surrounding the Cytokeratin that are left out by the mask. The current approach allows for capturing the Cytokeratin with all the surrounding DAPIs within the bounding box. MaskRCNN would not have masked any half-overlapping DAPIs. Through the black and white preprocessing we allow for variability in the incoming data.

The final box region loss is 0.0308542 and cross entropy loss for classification is 0.0023945. The final learning rate post decay is 0.000125.

\subsubsection*{Mask R-CNN}
Mask R-CNN is a masking algorithm which belongs to the computer vision sub-field of deep learning. To understand the complexities of the Mask R-CNN, we need to first understand R-CNNs, Region Based Convolutional Neural Networks. The algorithm breaks down an input image to sub-regions, which then get evaluated independently of the other regions by a CNN. The backbone of the Mask R-CNN algorithm however is a Faster R-CNN, an advanced iteration of the previous algorithm. Instead of feeding each region proposal (ROI: Region of Interest) to the convolution neural network, Faster R-CNN uses feature maps to summarize each ROI. RolPooling which stands for Region of Interest Pooling, used by the network allows for extracting small feature maps for each ROI.

Mask R-CNN is a network built on top of Faster R-CNN, and specializes in image segmentation tasks. There are 2 main types of image segmentation tasks: (1) semantic segmentation; (2) instance segmentation. 

\begin{figure}[htp]
    \centering
    \includegraphics[width=8cm]{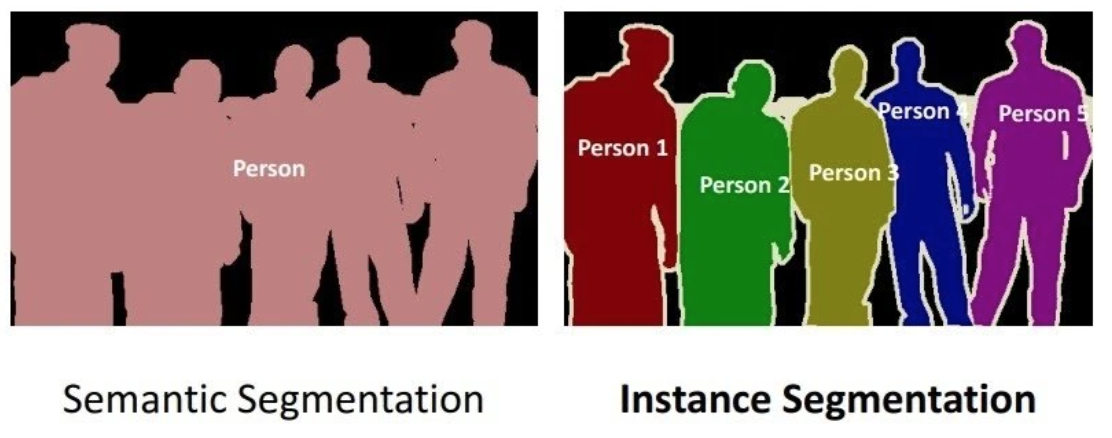}
    \caption{Different Segmentation Tasks}
    \label{fig:galaxy}
\end{figure}

For the task of identifying DAPI cells in the Multi-channel Enriched images, we needed to perform a meticulous instance segmentation task. We needed each DAPI to be independently identified, with clear boundaries around them. The algorithm therefore is of highest computer vision complexity, performing the combination of object detection, object localization, and object classification. 

Mask R-CNN adds a third output (or a second computation layer) to the final Faster-RCNN network, a mask for the object of desire, the objet petit a, which runs in parallel with the existing branch for bounding box recognition. Mask R-CNN is famous for its low marginal cost (on top of the Faster RCNN), and it's pixel to pixel alignment which the previously mentioned network lack. 

The configuration for our Mask R-CNN is:
\begin{enumerate}
    \item Batch Size: 16
    \item Base Learning Rate: 0.02
    \item Steps: (60000, 80000)
    \item Max Iter: 90000
    \item Number of Residual blocks [depth]: 50
\end{enumerate}

The cross-entropy loss for the model post-training on our 46 data points are 0.133003, with the mask region loss of 0.164770. The learning rate post-decay is 0.000125, with the Mask R-CNN classification accuracy of 0.93367. Total percentage of False Negatives in the validation set being 0.040939, and False Positive rate being 0.109557. The total loss for the overall model is 0.620905.

\end{document}